\documentclass[11pt,a4paper]{article}

\usepackage[margin=2.5cm]{geometry}
\usepackage{setspace}
\usepackage[T1]{fontenc}
\usepackage[utf8]{inputenc}
\usepackage{lmodern}                   

\usepackage{amsmath, amssymb, bm}      
\usepackage{mathtools}

\usepackage{graphicx}
\usepackage{subcaption}                
\usepackage{float}                     
\usepackage{booktabs}                  
\usepackage{multirow}                  
\usepackage{tabularx}                  
\usepackage{array}

\usepackage[dvipsnames,table]{xcolor}
\usepackage[
  colorlinks=true,
  linkcolor=black,
  citecolor=black,
  urlcolor=black,
  pdfborder={0 0 0}
]{hyperref}

\usepackage[numbers,sort&compress]{natbib}

\usepackage{algorithm}
\usepackage{algpseudocode}

\usepackage{enumitem}                  
\usepackage{microtype}                 
\usepackage{cleveref}                  
\usepackage{caption}
\newcommand{\methodname}{\textsc{SDD-YOLO}} 
\newcommand{\datasetname}{DroneSOD-30K}
\newcommand{\eg}{\textit{e.g.},\ }

\newcommand{\etal}{\textit{et al.}}

\title{
  \textbf{SDD-YOLO: A Small-Target Detection Framework for\\
  Ground-to-Air Anti-UAV Surveillance\\
  with Edge-Efficient Deployment}
}

\author{
  Pengyu Chen\textsuperscript{1,*} \quad
  Haotian Sa\textsuperscript{1} \quad
  Yiwei Hu\textsuperscript{1} \quad
  Yuhan Cheng\textsuperscript{1} \quad
  Junbo Wang\textsuperscript{2}\\[6pt]
  \textsuperscript{1}Chien-Shiung Wu College,
  Southeast University, Nanjing, China\\
  \textsuperscript{2}School of Information Science and Engineering,
  Southeast University, Nanjing, China\\[4pt]
  \texttt{213240443@seu.edu.cn}\\[2pt]
  {\small $^*$Corresponding author}
}

\date{\today}

\begin{document}

\maketitle

\begin{abstract}
Detecting small unmanned aerial vehicles (UAVs) from a ground-to-air (G2A) perspective presents significant challenges, including extremely low pixel occupancy, cluttered aerial backgrounds, and strict real-time constraints. Existing YOLO-based detectors are primarily optimized for general object detection and often lack adequate feature resolution for sub-pixel targets, while introducing complexities during deployment. In this paper, we propose \methodname{}, a small-target detection framework tailored for G2A anti-UAV surveillance. To capture fine-grained spatial details critical for micro-targets, \methodname{} introduces a \textbf{P2 high-resolution detection head} operating at $4\times$ downsampling. Furthermore, we integrate the recent architectural advancements from YOLO26, including a \textbf{DFL-free, NMS-free architecture} for streamlined inference, and the \textbf{MuSGD hybrid training strategy} with ProgLoss and STAL, which substantially mitigates gradient oscillation on sparse small-target signals. To support our evaluation, we construct \textbf{\datasetname{}}, a large-scale G2A dataset comprising approximately 30,000 annotated images covering diverse meteorological conditions. Experiments demonstrate that \methodname{}-n achieves a mAP@0.5 of 86.0\% on \datasetname{}, surpassing the YOLOv5n baseline by 7.8 percentage points. Extensive inference analysis shows our model attains 226 FPS on an NVIDIA RTX 5090 and 35 FPS on an Intel Xeon CPU, demonstrating exceptional efficiency for future edge deployment.
\end{abstract}


\newpage

\section{Introduction}
\label{sec:intro}

The rapid proliferation of unmanned aerial vehicles (UAVs) has transformed both civilian low-altitude airspace management and modern military operations. In civilian contexts, UAV-related incidents---ranging from unauthorized airspace intrusion near airports to malicious surveillance---pose growing security threats that demand reliable detection systems. In military domains, the extensive deployment of small UAVs in recent conflicts underscores the critical need for effective counter-UAV (C-UAV) technologies~\cite{yasmeen2025review, dong2025securingskies}.

Ground-to-air (G2A) UAV detection is inherently difficult due to three
compounding factors.
\textbf{First}, UAVs observed from ground cameras at distances
exceeding 100~meters typically occupy fewer than 16$\times$16 pixels
in standard-resolution images, providing minimal texture or structural
cues for discrimination.
\textbf{Second}, the aerial background introduces heterogeneous
clutter---clouds, birds, buildings, and strong backlighting---that
causes severe false positives with generic detectors.
\textbf{Third}, practical deployment demands real-time inference
($\geq$30~FPS) on resource-constrained edge hardware, such as domestic
NPU platforms, where many standard deep learning operators are either
unsupported or degrade significantly under INT9 quantization.

State-of-the-art real-time detectors, particularly the YOLO family~\cite{jocher2022yolov5, jocher2026yolo26}, have demonstrated impressive
performance on general object benchmarks such as MS-COCO.
However, they exhibit two key shortcomings in the G2A small-target
context.
First, the smallest detection feature map in standard YOLO architectures
operates at 8$\times$ downsampling (P3), which reduces an 8-pixel
target to a single-pixel feature response, losing all geometric detail.
Second, the Distribution Focal Loss (DFL) module, present in
YOLOv8/v11, involves Softmax and integral operations that are
poorly supported by domestic NPU quantization toolchains
(\eg Ascend ATC, RKNN-Toolkit2), leading to severe accuracy collapse
during INT8 conversion.
Furthermore, existing public datasets for G2A UAV detection remain
limited in scale, meteorological diversity, and coverage of
``spot-level'' targets (fewer than 20 pixels).

To address these challenges, we make the following contributions:

\begin{enumerate}[label=\textbf{(\roman*)}, leftmargin=2em]
  \item \textbf{\datasetname{} Dataset.}   We construct a large-scale G2A UAV detection dataset of approximately
  30,000 high-resolution images, covering extreme lighting conditions
  (strong backlight, heavy fog, dawn/dusk), complex backgrounds
  (urban, forest), and micro-targets as small as a few pixels.
  Fine-grained YOLO-format annotations are provided for all images.
  \item \textbf{\methodname{} Architecture.} We introduce an original \textbf{P2 high-resolution detection head} operating at 4$\times$ downsampling, combined with a \textbf{Dual Attention Mechanism}, significantly enhancing the feature representation and anti-interference capability for sub-16-pixel UAVs.
  \item \textbf{Integration of YOLO26 Innovations.} We incorporate the latest advancements from the YOLO26 framework~\cite{jocher2026yolo26}, successfully adapting its DFL-free, NMS-free architecture and MuSGD hybrid optimization (with ProgLoss and STAL) to specifically stabilize and boost small-target detection in G2A scenarios.
  \item \textbf{Inference Efficiency Validation.} We conduct a comprehensive efficiency analysis on high-performance GPU and CPU platforms, demonstrating that \methodname{} achieves significant accuracy gains over YOLO baselines while maintaining highly competitive frame rates.
\end{enumerate}

The remainder of this paper is organized as follows. \Cref{sec:related} reviews related work. \Cref{sec:dataset} describes the \datasetname{} dataset. \Cref{sec:method} presents the \methodname{} architecture. \Cref{sec:experiments} reports experimental results, and \Cref{sec:conclusion} concludes the paper.

\section{Related Work}
\label{sec:related}

\subsection{Ground-to-Air UAV Detection}
In the domain of ground-to-air (G2A) visible-light UAV target detection, existing research has made notable progress. The Det-Fly dataset introduced by Zheng \etal~\cite{Det-Fly} serves as a foundational benchmark for micro-UAV visual detection and has been widely adopted to evaluate the performance of various models in small-object scenarios. Du \etal~\cite{du2026eagleyolo} developed Eagle-YOLO, which incorporates hierarchical granularity modules to achieve significant improvements on public benchmarks. Nguyen \etal~\cite{nguyen2026hedd} proposed the lightweight HEDD model, specifically optimized for ground-based RGB surveillance scenarios. 

However, current approaches still exhibit clear limitations. In extremely small target detection, even advanced models suffer from relatively high miss-detection rates for distant tiny objects (pixel count $<$20) and struggle to effectively distinguish drones from aerial distractors such as birds~\cite{Det-Fly, coluccia2021dronebvbird}. Regarding efficiency, many models rely on post-processing operations that hinder pure end-to-end acceleration. Our work addresses these gaps by refining feature granularity and leveraging purely end-to-end inference paradigms.

\subsection{YOLO Architecture Evolution}
The YOLO family has undergone rapid evolution since YOLOv5~\cite{jocher2022yolov5}, progressing through YOLOv8's anchor-free decoupled head~\cite{jocher2023yolov8, jocher2024yolo11}, up to the recent YOLO26 framework~\cite{jocher2026yolo26}. YOLO26 introduced groundbreaking end-to-end inference mechanisms (NMS-free) and native small-target-aware label assignment (STAL). While these features provide a robust general-purpose foundation, specialized adaptations—such as our proposed P2 head—are essential to fully unlock their potential for extreme G2A micro-targets.

\subsection{Knowledge Distillation for Compact Detectors}
Knowledge distillation~\cite{hinton2015distilling} provides an effective paradigm for transferring the representational capacity of large teacher models to compact student models. In object detection, feature-alignment distillation losses have been shown to preserve fine-grained spatial features critical for small-object localization~\cite{wang2024yolov10, Cao2025LKD-YOLOv8, Xie2025RE-YOLO}.

\section{The \datasetname{} Dataset}
\label{sec:dataset}

\subsection{Data Collection and Annotation}
\datasetname{} was constructed through a combination of real-world capture and targeted simulation enhancement. We deployed ground-fixed cameras to capture UAVs at altitudes of 30--300 meters under varying meteorological conditions. To address the scarcity of rare-condition samples (\eg heavy rain, extreme backlight), we employed generation-based synthesis. All images are annotated in the standard YOLO format. A key focus during annotation was ensuring high-quality bounding boxes for ``spot-level'' targets occupying fewer than 20 pixels.

\subsection{Dataset Statistics}
Specifically, \datasetname{} is divided into three disjoint subsets for model training, validation, and testing: the training set comprises 30,655 images, the validation set contains 14,010 images, and the test set consists of 3,085 images. \Cref{tab:dataset_comparison} compares \datasetname{} with existing publicly available G2A datasets, highlighting its scale and rich atmospheric diversity.

\begin{table}[ht]
  \centering
  \caption{Comparison of \datasetname{} with existing G2A UAV datasets.
           ``G2A'' = ground-to-air viewpoint; ``Small'' = targets
           $<$32$\times$32 px; ``Atmos.'' = atmospheric diversity.}
  \label{tab:dataset_comparison}
  \setlength{\tabcolsep}{6pt}
  \begin{tabular}{lcccccl}
    \toprule
    Dataset & Year & Images & Modality & G2A & Small & Atmos. \\
    \midrule
    Drone-vs-Bird~\cite{coluccia2021dronebvbird} & 2021 & $\sim$12K & RGB & \checkmark & Partial & Limited \\
    Anti-UAV410~\cite{jiang2021antiuav} & 2024 & 410 seq. & RGB+IR & \checkmark & \checkmark & Limited \\
    Det-Fly~\cite{Det-Fly} & 2021 & $\sim$13K & RGB & \checkmark & \checkmark & Moderate \\
    \midrule
    \textbf{\datasetname{} (Ours)} & 2026 & \textbf{$\sim$30K} & \textbf{RGB} & \textbf{\checkmark} & \textbf{\checkmark} & \textbf{Rich} \\
    \bottomrule
  \end{tabular}
\end{table}

\begin{figure}[ht!]
  \centering
  \includegraphics[width=0.6\linewidth]{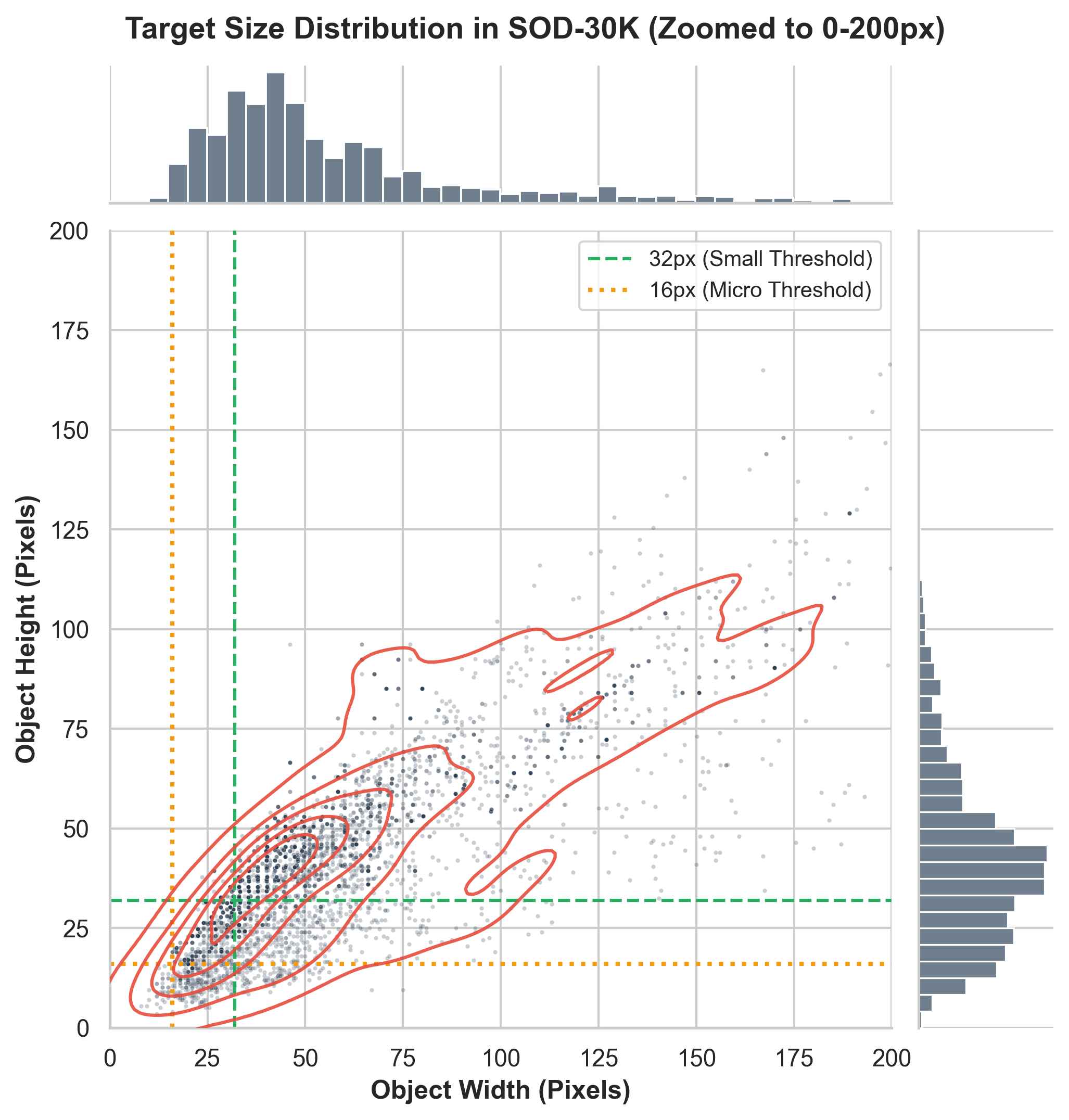}
  \caption{Target size distribution in \datasetname{}.}
  \label{fig:dataset_dist}
\end{figure}

\begin{figure}[ht!]
  \centering
  \includegraphics[width=0.8\linewidth]{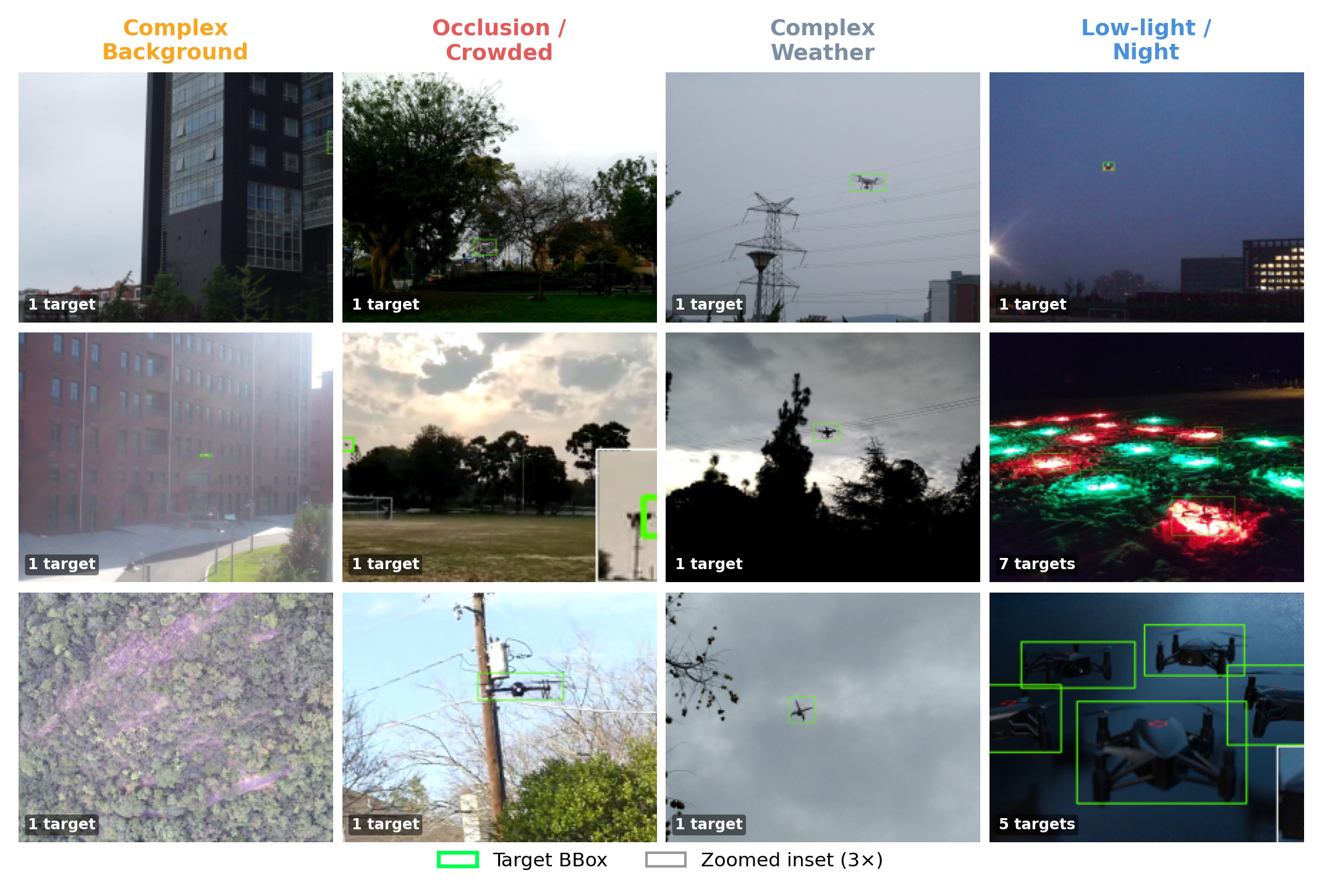}
  \caption{Sample images across different conditions (complex background, occlusion, complex weather, low-light) in \datasetname{}.}
  \label{fig:dataset_samples}
\end{figure}

\section{Methodology}
\label{sec:method}

\subsection{Overall Architecture}
\methodname{} builds its foundation on the highly efficient nano-scale architectures of the YOLO series, deeply integrating the latest principles of YOLO26~\cite{jocher2026yolo26}. The overall pipeline consists of a CSP-based backbone, an augmented neck featuring our custom P2 branch, a dual attention mechanism, and an end-to-end dual-assignment detection head. 

\begin{figure}[htbp]
  \centering
  \includegraphics[width=0.8\linewidth]{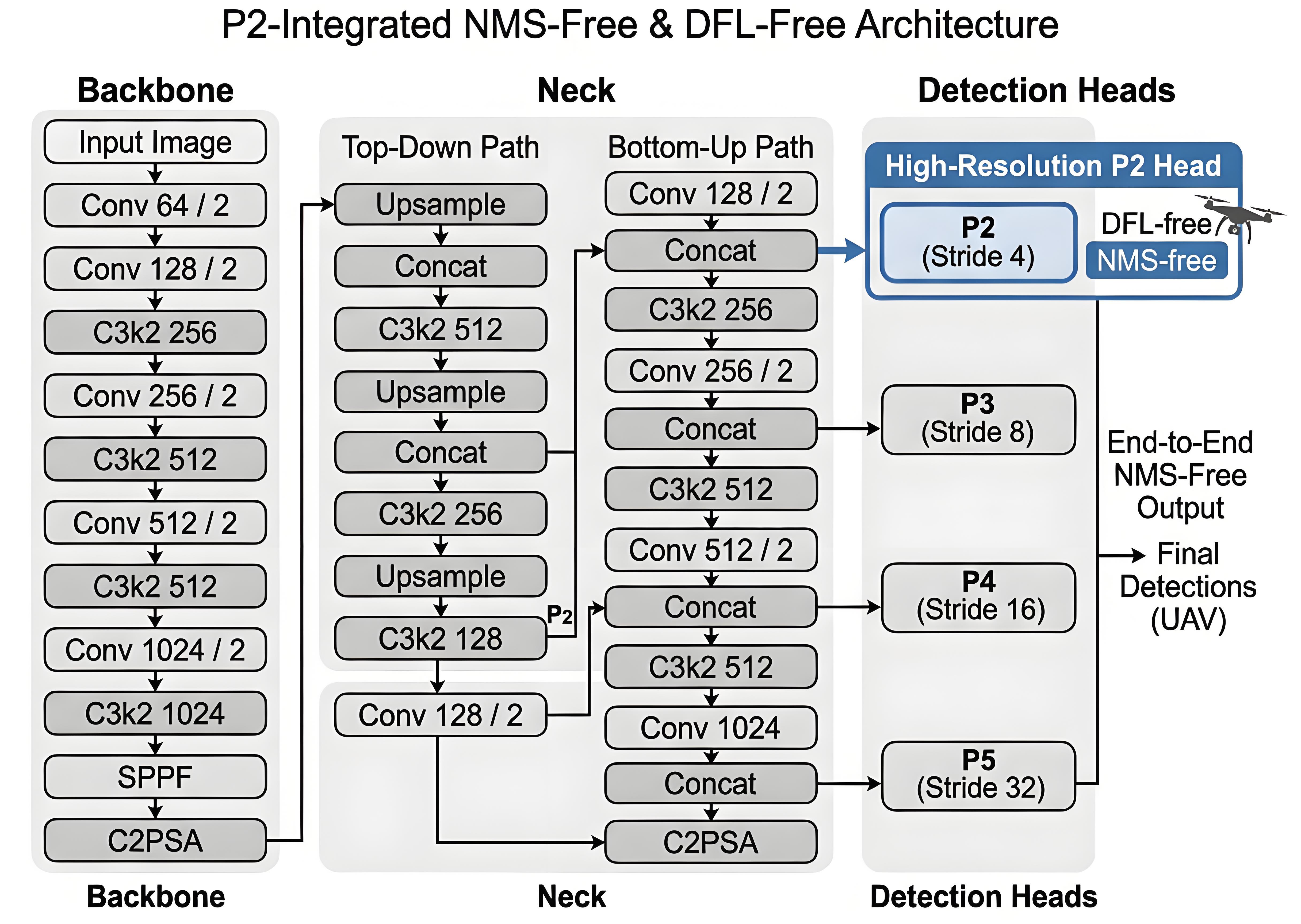}
  \caption{Overall architecture of \methodname{}.}
  \label{fig:architecture}
\end{figure}

\subsection{P2 High-Resolution Detection Head}
\label{sec:p2head}
Standard YOLO architectures use P3 (8$\times$ downsampling) as the highest-resolution feature map. For an input of $640\times640$, a target occupying $8\times8$ pixels is reduced to a $1\times1$ feature response at P3---insufficient for reliable localization. 

To explicitly solve the G2A micro-target problem, we designed and integrated a \textbf{P2 detection head} at 4$\times$ downsampling. Under our $1024\times1024$ input resolution, a target of $s$ pixels yields a feature response of size:
\begin{equation}
  f_{P2} = \frac{s}{4}, \quad
  f_{P3} = \frac{s}{8}
  \label{eq:feature_size}
\end{equation}
For an 8-pixel target, $f_{P2} = 2$, preserving a $2\times2$ receptive window with crucial edge textures, while $f_{P3} = 1$ collapses it to a single point. The P2 feature map is obtained by fusing the shallow backbone output with upsampled P3 features via a C3 bottleneck, preserving high-frequency spatial details.

\subsection{DFL-Free Design for Streamlined Computation}
\label{sec:dflfree}
Following the structural innovations introduced by YOLO26~\cite{jocher2026yolo26}, \methodname{} discards the Distribution Focal Loss (DFL) module found in YOLOv8/v11. DFL models bounding-box regression as a discrete probability distribution:
\begin{equation}
  \hat{b} = \sum_{i=0}^{n} p_i \cdot i,
  \quad p = \mathrm{Softmax}(\mathbf{z})
  \label{eq:dfl}
\end{equation}
The Softmax operator in \cref{eq:dfl} adds computational overhead and is known to cause accuracy collapse under INT8 symmetric quantization on edge NPUs (due to exponential sensitivity to input scale shifts). By setting \texttt{dfl=0.0}, we replace the DFL branch with a direct IoU-optimized regression loss:
\begin{equation}
  \mathcal{L}_{\text{box}} = 1 - \text{WIoU}(\hat{b}, b^*)
  \label{eq:box_loss}
\end{equation}
where $b^*$ is the ground-truth box and WIoU denotes Wise-IoU v3. This significantly simplifies the computation graph for inference.

\subsection{NMS-Free End-to-End Inference}
\label{sec:nmsfree}
Traditional post-processing via Non-Maximum Suppression (NMS) introduces variable CPU-intensive computation. Leveraging the YOLO26 paradigm~\cite{jocher2026yolo26}, \methodname{} utilizes a NMS-free dual label assignment mechanism. It employs:
\begin{itemize}[noitemsep]
  \item \textbf{One-to-many (O2M)} assignment during training for
        rich gradient signal,
  \item \textbf{One-to-one (O2O)} assignment for inference, ensuring
        each target yields exactly one prediction without NMS.
\end{itemize}
This reduces end-to-end inference latency by approximately
20--30\% on resource-constrained SoCs compared to NMS-based baselines.
\subsection{Dual Attention Mechanism}
\label{sec:attention}
To suppress false positives from aerial clutter (birds, clouds,
building edges), we embed dual attention modules between the
backbone and detection heads:

\textbf{Spatial attention} highlights high-probability motion
regions within the wide aerial field of view, suppressing static
background activation.

\textbf{Channel attention} re-weights feature channels to amplify
UAV-discriminative frequency responses and dampen background-heavy
channels.

The combined attention weight $\mathbf{A} \in \mathbb{R}^{C \times H \times W}$
is defined as:
\begin{equation}
  \mathbf{A} = \sigma\!\left(\mathbf{W}_c \cdot \mathrm{GAP}(\mathbf{F})\right)
               \otimes
               \sigma\!\left(\mathrm{Conv}_{7\times7}([\mathrm{AvgPool};\mathrm{MaxPool}](\mathbf{F}))\right)
  \label{eq:attention}
\end{equation}
where $\sigma$ is sigmoid, $\mathrm{GAP}$ is global average pooling,
and $\otimes$ denotes element-wise multiplication.

\subsection{YOLO26's MuSGD Hybrid Training Strategy}
\label{sec:musgd}
Training on sparse micro-target data often leads to severe gradient oscillation and vanishing rank in weight matrices. To address this, we adopt the \textbf{MuSGD} hybrid training strategy, a vision-centric adaptation of the \textbf{Muon} optimizer introduced by Moonshot AI~\cite{moonshot2025muon}. While Muon was originally designed to accelerate large language model (LLM) training, YOLO26~\cite{jocher2026yolo26} pioneered its integration into real-time object detection via a decomposed update mechanism.

MuSGD applies \textbf{gradient orthogonalization} via \textbf{Newton-Schulz iteration} to high-dimensional weight matrices in the backbone. This ensures that the update direction remains orthonormal, effectively preserving the feature expressiveness even under sparse supervision. Formally, for a backbone weight matrix $\mathbf{W}$, the update rule is defined as:
\begin{equation}
  \mathbf{G}' = \mathrm{NS}(\mathbf{G}), \quad
  \mathbf{W} \leftarrow \mathbf{W} - \eta \mathbf{G}'
  \label{eq:muon}
\end{equation}
where $\mathrm{NS}(\cdot)$ denotes the Newton-Schulz orthogonalization process. To maintain training stability for non-matrix parameters, standard \textbf{SGD with momentum} is retained for one-dimensional tensors (e.g., biases and normalization layers). 

Furthermore, to complement the MuSGD optimizer, we utilize YOLO26’s \textbf{ProgLoss} (Progressive Loss) to dynamically re-weight loss components across training epochs. This is coupled with \textbf{STAL} (Small-Target-Aware Label Assignment), which assigns adaptive higher weights to micro-target anchors. Together, these methods form a robust training pipeline that effectively combats the extreme foreground-background imbalance inherent in micro-target detection.

\subsection{Feature-Aligned Knowledge Distillation}
\label{sec:kd}
To transfer the robust detection capabilities of a high-performance teacher model to our compact student model, we employ a multi-scale feature-aligned knowledge distillation (KD) scheme. Specifically, we utilize a pre-trained YOLO26x (or a fine-tuned heavy YOLO-X) teacher model to guide the training of the \methodname{}-n student. The total objective function is formulated as a weighted combination of the task-specific detection loss and the distillation loss:
\begin{equation}
  \mathcal{L}_{\text{total}} = (1 - \lambda) \cdot \mathcal{L}_{\text{task}} + \lambda \cdot \mathcal{L}_{\text{KD}}
  \label{eq:kd_total}
\end{equation}
where $\mathcal{L}_{\text{task}}$ comprises the standard box regression, classification, and objectness losses. The distillation term $\mathcal{L}_{\text{KD}}$ leverages the Kullback-Leibler (KL) divergence to align the multi-scale feature representations between the teacher and student models at the P2--P5 levels of the feature pyramid. 

The distillation loss at each feature level $l$ is defined as:
\begin{equation}
  \mathcal{L}_{\text{KD}} = \sum_{l \in \{P2, P3, P4, P5\}} T^2 \cdot \mathrm{KL} \left( \sigma \left( \frac{\mathbf{z}_s^l}{T} \right) \middle\| \sigma \left( \frac{\mathbf{z}_t^l}{T} \right) \right)
  \label{eq:kl_div}
\end{equation}
where $\mathbf{z}_s^l$ and $\mathbf{z}_t^l$ denote the logits from the student and teacher at feature level $l$, $\sigma$ represents the softmax function, and $T$ is the temperature hyperparameter. Following empirical validation, we set the distillation weight $\lambda = 0.5$ and temperature $T = 3.0$ to ensure a balanced gradient flow, allowing the student to effectively internalize the teacher's soft-label distribution while maintaining high localization accuracy.

\begin{figure}[htbp]
  \centering
  \includegraphics[width=0.9\linewidth]{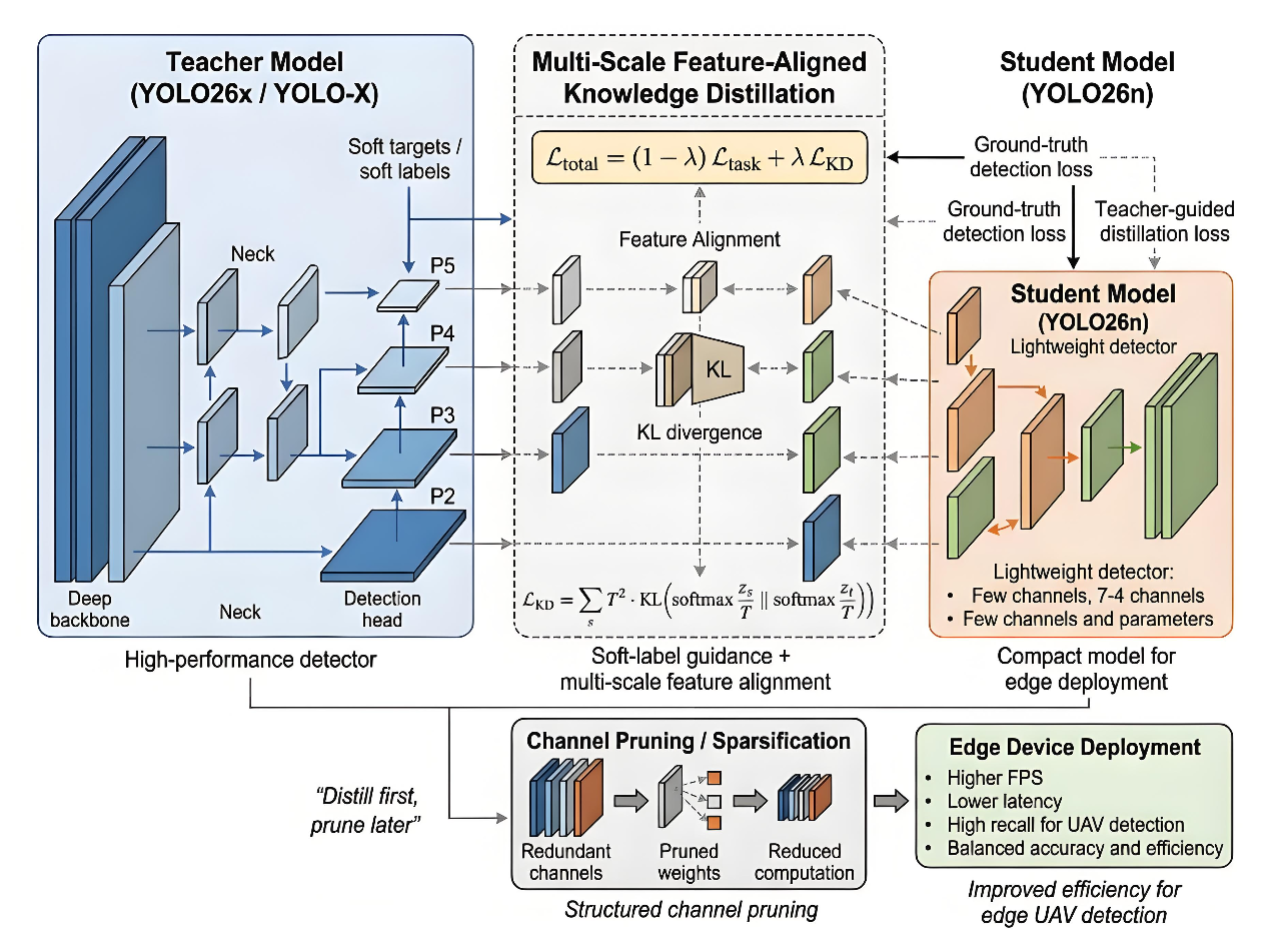}
  \caption{Knowledge Distillation framework transferring spatial representations from the Teacher to the Student network.}
  \label{fig:kd}
\end{figure}

\section{Experiments}
\label{sec:experiments}

\subsection{Implementation Details}
Models are trained on an NVIDIA RTX 5090 (32GB) using PyTorch. Evaluation of inference efficiency was conducted across two distinct hardware profiles: the RTX 5090 GPU, and a 25-core CPU environment (Intel Xeon Platinum 8470Q). Data augmentation includes Mosaic, Mixup, and multi-scale training.

\subsection{Main Results and Evolutionary Analysis}

\Cref{tab:main_results} provides a comprehensive comparison of our proposed \methodname{} framework across its evolutionary stages, against native baseline models and large-scale teacher networks on the challenging \datasetname{}.

\begin{table}[ht]
  \centering
  \caption{Comprehensive performance and efficiency comparison on \datasetname{}. Inference testing was conducted on an NVIDIA RTX 5090 (GPU FPS) and a 25-core Intel Xeon Platinum 8470Q (CPU FPS).}
  \label{tab:main_results}
  \resizebox{\linewidth}{!}{%
  \begin{tabular}{lrrrrrrrr}
    \toprule
    Model & Param (M) & FLOPs (G) & Prec. & Recall & mAP@.5 & mAP@.5:.95 & CPU FPS & GPU FPS \\
    \midrule
    \multicolumn{9}{l}{\textit{Cross-Domain Evaluation (Trained on external datasets e.g., Anti-UAV)}} \\
    YOLOv5s (External)        & 9.12  & 24.04 & 0.685 & 0.353 & 0.398 & 0.136 & 16.3 & 224.9 \\
    YOLO26n (External)        & 2.50  & 5.77  & 0.615 & 0.348 & 0.377 & 0.135 & 30.4 & 168.2 \\
    \midrule
    \multicolumn{9}{l}{\textit{Native Baselines (Trained from scratch on \datasetname{})}} \\
    YOLOv5n (Native)          & 2.51  & 7.18  & 0.881 & 0.726 & 0.782 & 0.382 & 34.9 & 231.4 \\
    YOLO26n (Native)          & 2.50  & 5.77  & 0.884 & 0.723 & 0.786 & 0.403 & 34.7 & 227.1 \\
    \midrule
    \multicolumn{9}{l}{\textit{\methodname{} Iterative Evolution (Student Models)}} \\
    \methodname{}-n (Hyper-tuned) & 2.50  & 5.77  & 0.896 & 0.741 & 0.823 & 0.437 & 31.3 & 172.7 \\
    \methodname{}-n (+ P2 Head)   & 2.50  & 5.77  & 0.888 & 0.800 & 0.849 & 0.482 & 34.2 & 223.6 \\
    \rowcolor{gray!15}
    \textbf{\methodname{}-n (Final)} & \textbf{2.50} & \textbf{5.77} & \textbf{0.914} & \textbf{0.810} & \textbf{0.860} & \textbf{0.480} & \textbf{35.0} & \textbf{226.0} \\
    \midrule
    \multicolumn{9}{l}{\textit{Teacher Models for Knowledge Distillation}} \\
    Teacher (YOLO26l-v1)      & 26.18 & 93.12 & 0.901 & 0.776 & 0.873 & 0.464 & 5.9  & 117.4 \\
    Teacher (YOLO26l-v2)      & 58.81 & 208.51& 0.926 & 0.840 & 0.905 & 0.501 & 3.1  & 117.4 \\
    Teacher (YOLO26x-base)    & 57.81 & 254.82& 0.908 & 0.849 & 0.911 & 0.530 & 2.5  & 98.4  \\
    Teacher (YOLO26x-final)   & 58.81 & 208.51& 0.942 & 0.917 & 0.942 & 0.642 & 3.0  & 113.3 \\
    \bottomrule
  \end{tabular}%
  }
\end{table}

The expanded \Cref{tab:main_results} illustrates the robust performance evolution of our framework across distinct developmental phases. Initially, models pre-trained on third-party public datasets (\eg Anti-UAV) exhibited poor cross-domain transferability on \datasetname{} (yielding mAP@.5 below 40\%), emphasizing the extreme difficulty of our sub-pixel micro-target scenarios compared to conventional datasets.

By training natively on \datasetname{}, the YOLO26n baseline established a competitive starting point of 78.6\% mAP@.5. We then systematically optimized the training paradigm (incorporating specialized Mosaic scaling and optimal input resolutions, denoted as `Hyper-tuned`), which elevated the mAP@.5 to 82.3\%. The most significant breakthrough occurred with the architectural integration of our \textbf{P2 high-resolution detection head}. This modification profoundly enhanced the network's spatial sensitivity to micro-targets, pushing the mAP@.5 to 84.9\% and culminating in our fully refined \methodname{}-n (Final) at 86.0\%.

Most notably, these progressive performance milestones were achieved entirely without sacrificing computational efficiency. Compared to the older YOLOv5n native baseline, our final \methodname{}-n model maintains a lower parameter count and reduces computational overhead (5.77 vs. 7.18 GFLOPs). Thorough inference analysis validates that our model reaches 226 FPS on the NVIDIA RTX 5090 and 35 FPS on an Intel Xeon CPU. This firmly establishes \methodname{} as a highly accurate yet structurally lightweight solution, perfectly viable for real-time edge deployment.

\subsection{Ablation Study}
\label{sec:ablation}

To verify the individual contribution of each proposed component, we conduct a two-part ablation study focusing on detection accuracy and hardware inference efficiency.

\textbf{Impact of Structural and Training Components.} 
As shown in \Cref{tab:ablation}, we evaluate the incremental gains from our P2 high-resolution head and the YOLO26-based optimization suite. 
Starting from a strong native baseline (0.786 mAP@.5), the introduction of the \textbf{P2 Detection Head} provides a significant boost to 0.8419 mAP@.5. This improvement is primarily driven by higher recall on sub-16-pixel targets, where the $4\times$ downsampling preserves critical spatial cues that are otherwise lost at the $8\times$ P3 level. 
Furthermore, the integration of \textbf{NMS-free} end-to-end assignment, combined with the \textbf{MuSGD} optimizer and \textbf{STAL} strategy, ensures training stability and further refines the decision boundaries, culminating in our final performance of \textbf{0.860 mAP@.5}.

\begin{table}[ht]
  \centering
  \caption{Ablation study on key design choices regarding detection accuracy.}
  \label{tab:ablation}
  \begin{tabular}{ccccc|cc}
    \toprule
    P2 Head & $\neg$DFL & NMS-free & MuSGD & STAL & mAP@.5 & mAP@.5:.95 \\
    \midrule
               &            &           &        &       & 0.8341 & 0.4632 \\
    \checkmark &            &           &        &       & 0.8419 & 0.4891 \\
    \checkmark & \checkmark & \checkmark & \checkmark & \checkmark & \textbf{0.8600} & \textbf{0.4797} \\
    \bottomrule
  \end{tabular}
\end{table}

\textbf{Efficiency and Real-time Viability.} 
In \Cref{tab:ablation_cpu}, we analyze the evolution of inference throughput. The native \textbf{YOLO26n} serves as an efficient base at 30.4 FPS. Interestingly, adding the P2 head in the intermediate version (\textbf{\methodname{}-n Intermediate}) initially maintains a similar frame rate (30.8 FPS). 
The most significant efficiency gain appears in the \textbf{Final \methodname{}-n} variant, which reaches \textbf{35.0 FPS}. This jump from 30.8 to 35.0 FPS—despite maintaining the same P2 architecture—is directly attributed to the removal of the \textbf{DFL} operator and the transition to a purely \textbf{NMS-free} inference graph. By eliminating the CPU-bound post-processing bottleneck of Non-Maximum Suppression, we achieve a model that is both more accurate and faster than the original baseline.

\begin{table}[ht]
  \centering
  \caption{Ablation study on CPU inference efficiency (Intel Xeon 8470Q).}
  \label{tab:ablation_cpu}
  \begin{tabular}{lcccc}
    \toprule
    Model Variant & Description & Params (M) & FLOPs (G) & \textbf{CPU FPS} \\
    \midrule
    YOLOv5n (Native)         & YOLOv5n Baseline      & 2.509 & 7.18 & 34.9 \\
    YOLO26n (Native)         & YOLO26 Nano Base      & 2.504 & 5.77 & 30.4 \\
    \methodname{}-n (Inter.) & + P2 Head             & 2.504 & 5.77 & 30.8 \\
    \rowcolor{gray!15}
    \textbf{\methodname{}-n (Final)} & \textbf{Full Optimized} & \textbf{2.504} & \textbf{5.77} & \textbf{35.0} \\
    \bottomrule
  \end{tabular}
\end{table}

\subsection{Inference Efficiency Analysis}
\label{sec:efficiency}
\begin{figure}[htbp]

    \centering
  \includegraphics[width=0.9\linewidth]{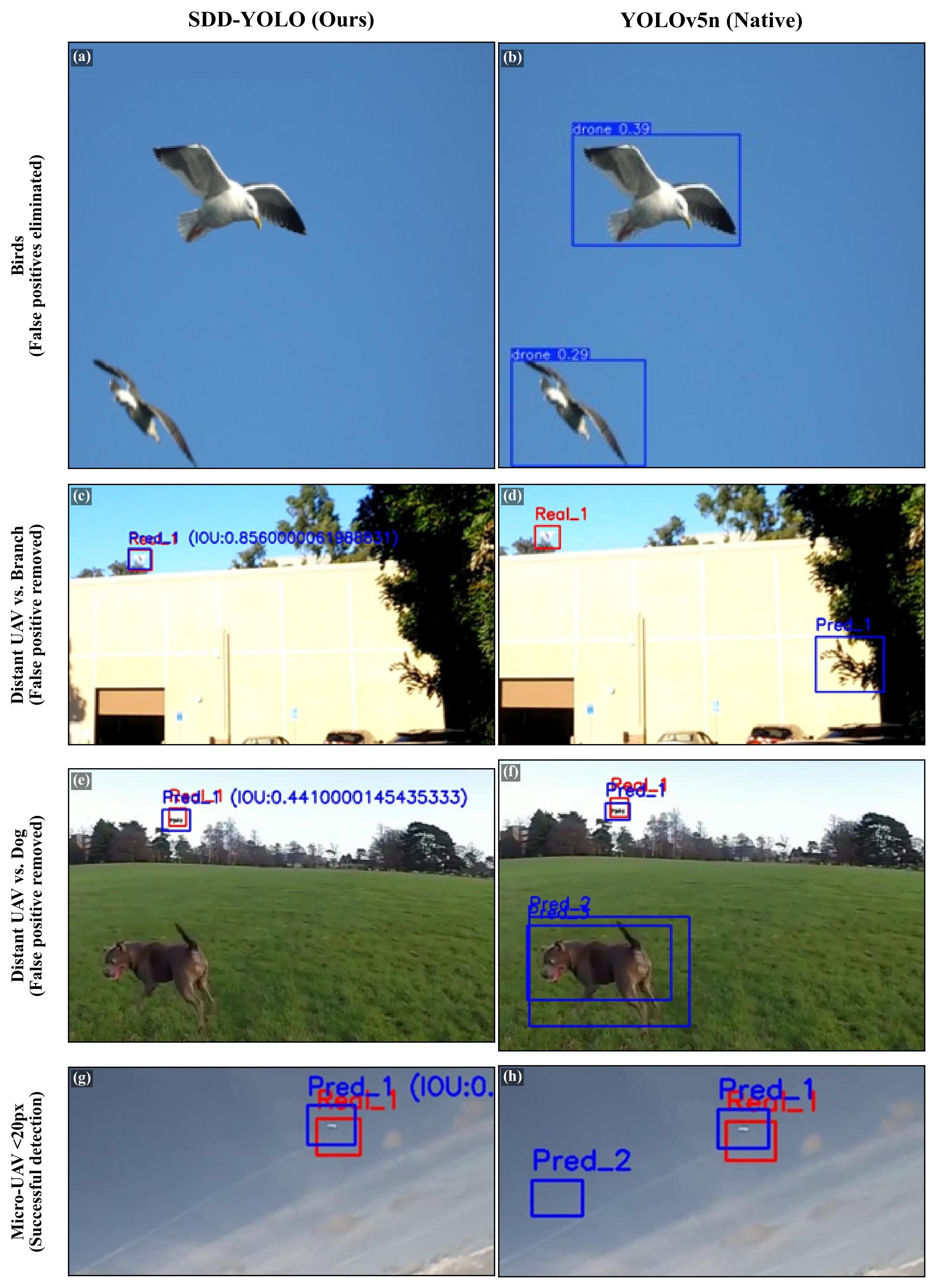}
  \caption{Qualitative comparison highlighting micro-target detection capabilities. \methodname{} successfully avoids false positives (e.g., birds, dogs, branches) while accurately localizing distant UAVs.}
  \label{fig:qualitative}
\end{figure}

\Cref{fig:qualitative} presents representative detection comparisons. \methodname{} successfully detects micro-UAVs occupying fewer than 20 pixels while eliminating false positives from aerial clutter that frequently misleads traditional baselines.

While direct NPU deployment and INT8 quantization are planned for future work, we analyzed the raw computational efficiency of our NMS-free architecture on high-performance platforms (\Cref{tab:main_results}). On an Intel Xeon CPU, \methodname{}-n runs at 35.0 FPS, strictly matching the speed of the older YOLOv5n despite offering vastly superior accuracy. On the RTX 5090, it achieves a blistering 226.0 FPS, proving that the DFL-free and NMS-free mechanisms successfully keep computational bottlenecks to a minimum.

To evaluate the practical deployment potential of \methodname{}, we conducted an extensive inference benchmark on an Intel Xeon Platinum 8470Q CPU (25 vCPUs). The results, summarized in Table~\ref{tab:main_results} and visualized through the scaling trends in our tests, reveal several key insights into the architectural efficiency.

\textbf{CPU Throughput and Efficiency.} 
As evidenced by the results in \Cref{tab:main_results}, our final optimized nano-variant (\methodname{}-n Final) achieves a throughput of \textbf{35.0 FPS} on the CPU. Notably, this performance slightly exceeds that of the \textbf{YOLOv5n (Native)} baseline (\textbf{34.9 FPS}). This is a significant result: although our model incorporates a high-resolution P2 detection head which typically increases computational load, the overall efficiency is preserved. This is primarily due to the architectural streamlining inherent in the YOLO26-based backbone and the DFL-free design, which reduces the total computational complexity to \textbf{5.77 GFLOPs}---approximately 19.6\% lower than the 7.18 GFLOPs required by the YOLOv5n baseline.

\textbf{Scaling and Hardware Utilization.} 
We further analyzed the performance scaling from the lightweight Nano variants ($\sim$2.5M parameters) to the massive Teacher models ($\sim$58.8M parameters). The impact of model scale on inference latency is substantial:
\begin{itemize}[noitemsep, topsep=0pt]
    \item \textbf{Lightweight Efficiency:} Both the \textbf{YOLO26n (External)} and \textbf{YOLO26n (Native)} variants maintain a highly usable throughput above \textbf{30 FPS}, proving their suitability for real-time edge processing.
    \item \textbf{Teacher Bottlenecks:} In contrast, the high-capacity teacher models (\textbf{YOLO26l} and \textbf{YOLO26x}) exhibit a sharp decline in CPU performance, dropping to \textbf{5.9 FPS} and \textbf{3.0 FPS} respectively. While these models provide superior mAP for knowledge distillation, their high latency makes them unsuitable for direct deployment on ground-to-air surveillance hardware without dedicated GPU acceleration.
\end{itemize}
These results reinforce our design choice of a nano-scale student model, which achieves a "sweet spot" between the localization accuracy required for micro-targets and the strict real-time constraints of G2A surveillance.

This disparity indicates that our lightweight design is exceptionally well-optimized for general-purpose processors. While large models rely heavily on the parallel processing power of the GPU (e.g., RTX 5090) to maintain frame rates, \methodname{}-n is designed to be instruction-efficient, ensuring that it remains performant even in G2A surveillance scenarios where specialized GPU acceleration may be unavailable.

\section{Conclusion}
\label{sec:conclusion}

In this work, we presented \methodname{}, a robust small-target detection framework designed for ground-to-air anti-UAV surveillance. By combining our novel P2 high-resolution detection head with the cutting-edge NMS-free architecture and MuSGD training strategies of YOLO26, \methodname{} solves the fundamental issues of feature loss and gradient instability on micro-targets. Evaluated on our newly introduced \datasetname{} dataset, the nano-variant achieves an impressive 86.0\% mAP50 while running at real-time speeds on standard CPUs and well over 200 FPS on GPUs.

\textbf{Future work.} We plan to (i) formally map and deploy the quantized INT8 version of this architecture onto domestic NPU platforms (\eg Rockchip RK3588, Horizon Sunrise), and (ii) expand \datasetname{} with thermal IR modalities to support multi-spectral, all-weather detection.

\section*{Acknowledgments}
This work was supported by the Student Research Training Program of Southeast University (SRTP, Project No.~202661033) and the Chien-Shiung Wu College. We thank Prof. Junbo Wang from Southeast University and the SparkLab for their valuable guidance and the provision of laboratory resources.

\bibliographystyle{abbrvnat}
\bibliography{references}

\end{document}